%% file: main.tex
\documentclass[10pt,twocolumn,letterpaper]{article}

\usepackage[pagenumbers]{iccv} 


\definecolor{iccvblue}{rgb}{0.21,0.49,0.74}
\usepackage[pagebackref,breaklinks,colorlinks,allcolors=iccvblue]{hyperref}

\usepackage{multirow}
\usepackage{comment}
\usepackage{bbding}

\newcommand{\Mark}[1]{\textsuperscript{#1}}

\newlength\savewidth

\title{Towards Scalable Video Anomaly Retrieval: A Synthetic Video-Text Benchmark}

\author{Shuyu Yang$^1$ \quad Yilun Wang$^1$ \quad Yaxiong Wang$^{2}$ \quad Li Zhu$^{1}$ \Mark{\Envelope} \quad Zhedong Zheng$^3$ \Mark{\Envelope}\\
$^1$Xi'an Jiaotong University, $^2$Hefei University of Technology, $^3$University of Macau\\
{\tt\small \{ysy653, wangyl, wangyx15\}@stu.xjtu.edu.cn, zhuli@mail.xjtu.edu.cn, zhedongzheng@um.edu.mo}
}

\begin{document}
 \input{sec/0_abstract}

 \input{sec/1_intro}

 \input{sec/2_related}
 \input{sec/3_bench}
 \input{sec/5_experiment}
 \input{sec/6_conclusion}
 {
     \small
     \bibliographystyle{ieeenat_fullname}
     \bibliography{main}
 }

\end{document}

%% file: sec/0_abstract.tex
\pagestyle{plain}
\twocolumn[{
\renewcommand\twocolumn[1][]{#1}
\maketitle
\begin{center}
\setlength{\abovecaptionskip}{8pt} 
\setlength{\belowcaptionskip}{0pt} 
\centering
\vspace{-.05in}
\includegraphics[height=1.8in, width=6.8in]{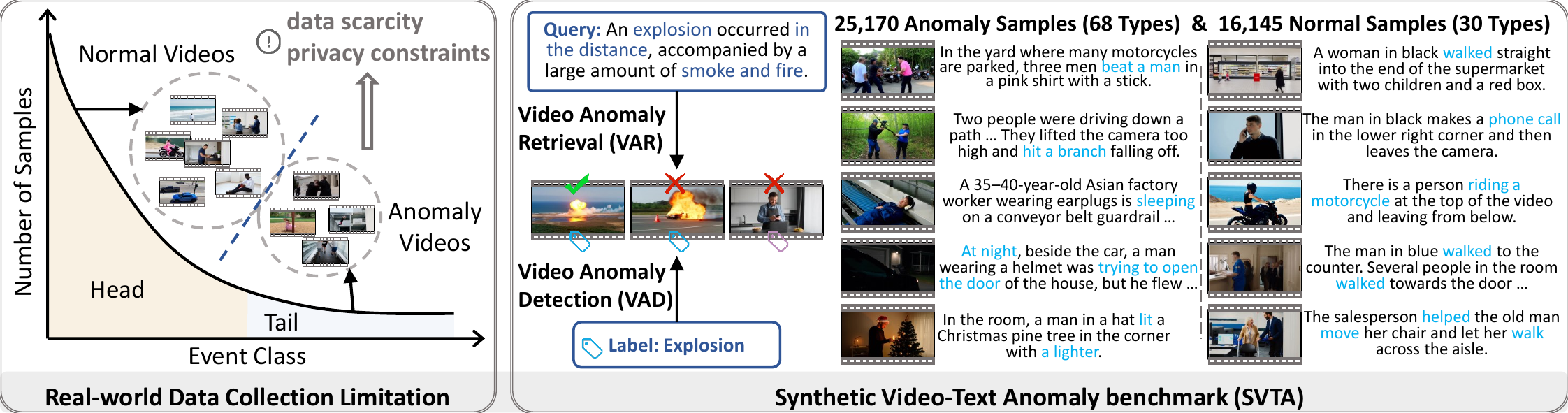}
\vspace{-.05in}
\captionof{figure}
{Due to the {\bf long-tail distribution} of real-world anomaly and {\bf privacy constraints}, {\bf large-scale data collection remains challenging (\emph{left})}. To address this limitation, we construct {\bf a large-scale Synthetic Video-Text Anomaly (SVTA) benchmark (\emph{right})}, which comprises {\bf 41,315 video-text pairs} covering {\bf 68 anomaly types} and {\bf 30 normal events}.
}
\label{fig:teaser}
\end{center}}]

\begin{abstract}
Video anomaly retrieval aims to localize anomalous events in videos using natural language queries to facilitate public safety.
However, existing datasets suffer from severe limitations: (1) data scarcity due to the long-tail nature of real-world anomalies, and (2) privacy constraints that impede large-scale collection. To address the aforementioned issues in one go, we introduce SVTA (Synthetic Video-Text Anomaly benchmark), the first large-scale dataset for cross-modal anomaly retrieval, leveraging generative models to overcome data availability challenges.
Specifically, we collect and generate video descriptions via the off-the-shelf LLM (Large Language Model) covering 68 anomaly categories, \eg, throwing, stealing, and shooting. These descriptions encompass common long-tail events. We adopt these texts to guide the video generative model to produce diverse and high-quality videos.
Finally, our SVTA involves 41,315 videos (1.36M frames) with paired captions, covering 30 normal activities, \eg, standing, walking, and sports, and 68 anomalous events, \eg, falling, fighting, theft, explosions, and natural disasters. 
We adopt three widely-used video-text retrieval baselines to comprehensively test our SVTA, revealing SVTA's challenging nature and its effectiveness in evaluating a robust cross-modal retrieval method. 
SVTA eliminates privacy risks associated with real-world anomaly collection while maintaining realistic scenarios.
The dataset demo is available at: [\url{https://svta-mm.github.io/SVTA.github.io/}].
\end{abstract}

%% file: sec/1_intro.tex
\section{Introduction}

\begin{table*}[t]
  \centering
  \vspace{-.15in}
  \resizebox{0.96\textwidth}{!}{
  \footnotesize
  \begin{tabular}{l|cccccccc}
    \toprule
    \multirow{1}{*}{Datasets} & Modality & \multirow{1}{*}{Annotation} & \multirow{1}{*}{Anno. Format} & \#Videos  & \multirow{1}{*}{\#Texts} & \multirow{1}{*}{\#Anomaly Types} & \multirow{1}{*}{Anomaly:Normal} & \multirow{1}{*}{Data source} \\
    \midrule
    UBnormal~\cite{acsintoae2022ubnormal} & Video & Frame-level Tag & Action Label & 543 & - & 22 Anomaly & 2:3  & Synthesis \\
    ShanghaiTech~\cite{luo2017revisit} & Video & Frame-level Tag & Action Label & 437 & - & 11 Anomaly & 1:18 & Collection \\
    UCF-Crime~\cite{sultani2018real} & Video & Video-level Tag & Action Label & 1,900 & - & 13 Anomaly & 1:1 & Collection \\
    UCA~\cite{yuan2024towards} & Video, Text & Video-level Text & Action Text & 1,900 & 23,542 & 13 Anomaly & 1:1 & Collection \\
    UCFCrime-AR~\cite{wu2024toward} & Video, Text & Video-level Text & Action Text & 1,900 & 1,900 & 13 Anomaly & 1:1 & Collection \\
    \hline
    SVTA (Ours) & Video, Text & Video-level Text & Action Text & \textbf{41,315} & \textbf{41,315} & 68 Anomaly & 3:2 & Synthesis \\
    \bottomrule
  \end{tabular}
  }
  \vspace{-.1in}
  \caption{{\bf Comparison of the proposed SVTA dataset and some of the other publicly available datasets for anomaly detection and anomaly retrieval.} Our dataset provides many more video samples, action classes (anomaly and normal), and background in comparison with other available datasets for anomaly retrieval (Anno. means Annotation).
  }
  \label{tab:dataset}
  \vspace{-.15in}
\end{table*}

Video Anomaly Detection (VAD)~\cite{cao2024context}, typically defined as a binary or multiclass classification task, holds significant value in applications such as hazard warning and intelligent security. Although existing deep learning-based methods~\cite{micorek24mulde} have achieved remarkable progress in VAD, single coarse-grained labels prove to be insufficient to describe the complexity of real-world events in practical applications~\cite{wu2024toward}. To address this, Wu \etal~\cite{wu2024toward} propose Video Anomaly Retrieval (VAR), which aims to retrieve relevant video segments from untrimmed long videos through cross-modal queries (text descriptions or synchronized audio). Compared to obtaining numerous video results using single labels, \eg, ``vandalism'', VAR allows users to retrieve video clips containing specific anomalous events using detailed textual queries, \eg, ``multiple intruders breaking into a house at night'', offering greater practical utility (as shown in Figure \ref{fig:teaser} (right)).

However, existing VAR benchmarks~\cite{wu2024toward, yuan2024towards} are primarily constructed by annotating mainstream VAD dataset (UCF-Crime~\cite{sultani2018real}) with text. Consequently, these VAR datasets inherit inherent limitations from their parent VAD datasets (as shown in Table \ref{tab:dataset}):
{\bf (1) Severe Data Scarcity:} Current anomaly datasets contain at most 1,900 video samples. Such limited data struggles to meet the generalization requirements of data-hungry deep learning methods.
{\bf (2) Restricted Anomaly Categories:} Most datasets cover only limited anomaly types (22 at most). However, real-world anomalies exhibit vast diversity and strong context dependence. Models performing well on fixed-category datasets may fail in complex real-world retrieval scenarios.
{\bf (3) Temporal Sparsity of Anomalies:} Anomalous events are highly sparse along video timelines, with most frames containing normal or irrelevant content. Accurately localizing and sampling anomalous frames remains a challenge.
{\bf (4) Long-tail Distribution:} Due to high data collection costs and the sporadic nature of anomalies (as shown in Figure \ref{fig:teaser} (left)), most datasets suffer from imbalanced category distributions.
{\bf (5) Ethical and Privacy Concerns:} Anomaly videos often involve human subjects. Current data collection practices lack adequate consideration of privacy protection and demographic biases.

To address these challenges, we propose a large-scale {\bf S}ynthetic {\bf V}ideo-{\bf T}ext {\bf A}nomaly Benchmark ({\bf SVTA}) for text-based video anomaly retrieval. SVTA contains {\bf 41,315} video-text pairs: texts describing events or behaviors are collected from off-the-shelf datasets or generated by LLMs, while videos are synthesized via text-guided generative models. Each video contains 33 frames (15 fps) with strategically concentrated keyframes. The dataset covers:
{\bf 30 normal activities: }Including typical daily actions, \eg, walking and standing; and context-compatible activities, \eg, playing soccer on a field.
{\bf 68 types of anomalies: } focusing on biomechanical deviations, \eg, falling and lying down; contextual violations, \eg, running in a laboratory; breaches of social norms, \eg, hitting and pushing; and environmental anomalies, \eg, natural disasters and objects falling.
In Figure \ref{fig:teaser} (right), we show some anomaly and normal samples of SVTA.
The normal-to-anomalous video ratio of 2:3 mitigates long-tail distribution issues in SVTA. To enhance realism and diversity, LLM-generated texts incorporate variations in human age, gender, and cultural backgrounds, environments, and camera perspectives. Furthermore, synthetic data collection bypasses privacy concerns while reducing potential biases. We evaluate three representative video-text retrieval methods, \ie, CLIP4Clip~\cite{luo2022clip4clip}, X-CLIP~\cite{ma2022x}, and GRAM~\cite{cicchetti2025gramian}, on SVTA, revealing the challenges posed by our dataset. Furthermore, we explore zero-shot performance on two real-world anomaly datasets (UCFCrime-AR~\cite{wu2024toward} and OOPS!~\cite{epstein2020oops}). Models trained on SVTA achieve competitive zero-shot retrieval performance on these benchmarks.
Our main contributions are summarized as follows:
\begin{itemize}[leftmargin=*]
\item We construct a large-scale {\bf S}ynthetic {\bf V}ideo-{\bf T}ext {\bf A}nomaly Benchmark ({\bf SVTA}) to deal with data scarcity in video anomaly research. SVTA contains 41K videos (1.36M frames) with paired captions, covering 30 normal and 68 anomalous event categories (normal-to-anomalous ratio: 2:3). The dataset features demographic and environmental diversity while addressing the privacy concerns and mitigating the long-tail biases.
\item We benchmark three prevailing video-text retrieval methods on SVTA. We observe that the model trained on our SVTA dataset shows competitive zero-shot generalization to two real-world datasets, including UCFCrime-AR and OOPS!.
\end{itemize}

\begin{figure*}[t]
  \centering
  \vspace{-.15in}
  \includegraphics[width=0.98\linewidth]{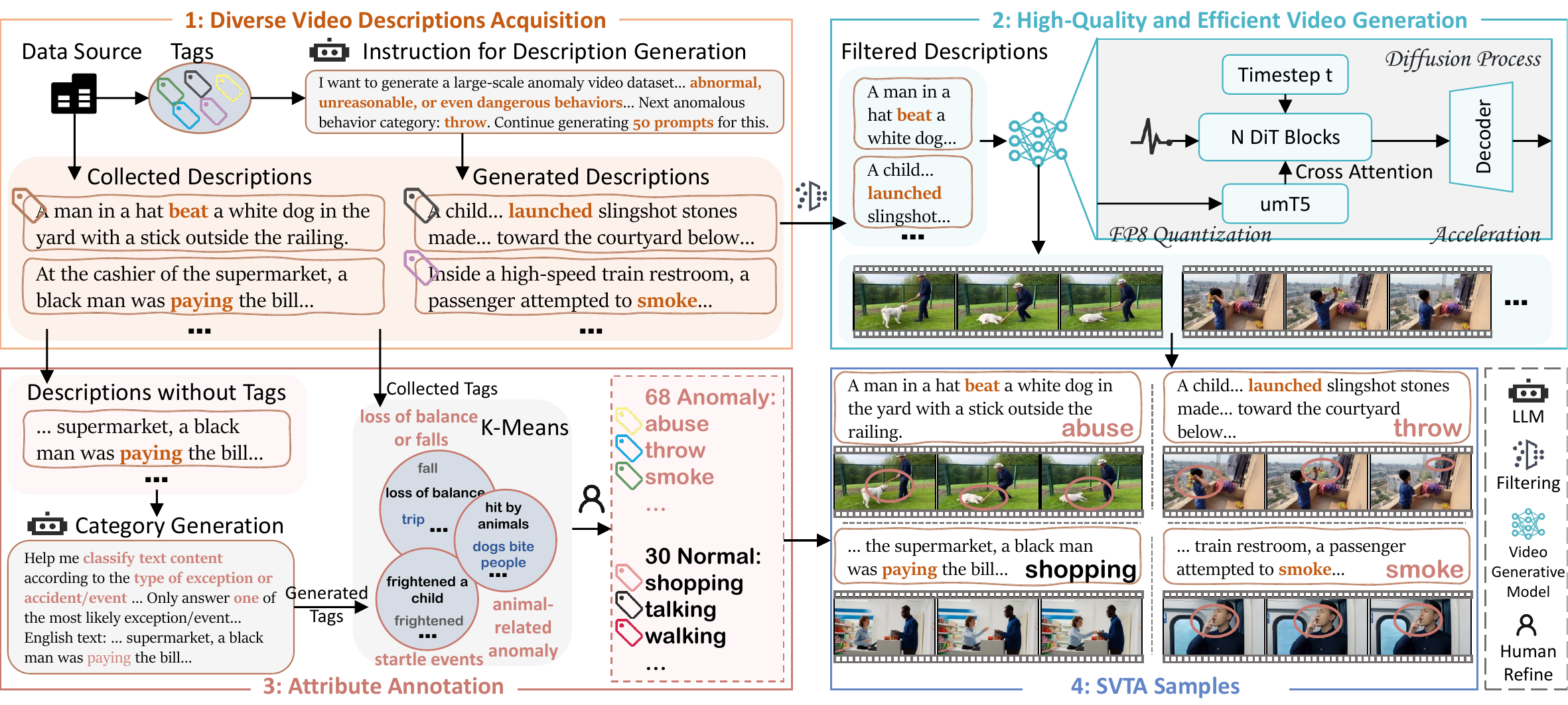}
  \vspace{-.1in}
  \caption{{\bf Pipeline of our Synthetic Video-Text Anomaly (SVTA) benchmark construction.} First, we collect and generate diverse video descriptions via LLM. Second, we leverage a state-of-the-art open-source video generative model to craft high-quality videos. Third, we adopt LLM to assign preliminary attributes for samples lacking explicit normal/anomaly labels and refine all labels by K-Means clustering and manual verification. The final dataset integrates 41,315 rigorously curated video-text pairs.}
  \label{fig:pipe}
  \vspace{-.15in}
\end{figure*}

%% file: sec/2_related.tex
\section{Related Work}
\noindent\textbf{Video Anomaly Detection.}
Video Anomaly Detection (VAD) aims to identify events in videos that deviate from normal patterns.
Significant progress has been made in recent years in deep learning-based VAD research. 
VAD typically involves various settings~\cite{micorek24mulde}: one-class classification (no abnormal data available during training)~\cite{Park2020, Sabokrou2017, Liu2018, Wu2019}, unsupervised learning (abnormalities present in the training set but unknown in which videos)~\cite{zaheer2022generative}, and supervised or weakly-supervised tasks (training labels indicate abnormal frames or videos containing anomalies)~\cite{Wu2021, wu2020not, sultani2018real, feng2021mist}.
Researchers have proposed different benchmarks tailored to these various settings. Almost all existing video datasets exhibit limitations regarding scene diversity, types of anomalies, and the number of videos. For instance, CUHK Avenue~\cite{lu2013abnormal}, Street Scene~\cite{ramachandra2020street}, Subway Entrance~\cite{adam2008robust}, Subway Exit~\cite{adam2008robust}, UCSD Ped1~\cite{li2013anomaly}, and UCSD Ped2~\cite{li2013anomaly} contain videos from only a single scene.
While the ShanghaiTech Campus dataset ~\cite{luo2017revisit} and the synthetic UBnormal dataset ~\cite{acsintoae2022ubnormal} cover 13 and 29 distinct scenes, respectively, they still struggle to match the diversity of real-world scenarios. 
UCF-Crime ~\cite{sultani2018real} encompasses multiple scenes, but is limited to only 13 types of real-world anomalous events, such as Abuse and Burglary. 
Furthermore, all the aforementioned datasets are constrained by the number of videos; even UCF-Crime ~\cite{sultani2018real}, which has the largest volume, contains only 1,900 videos.
Building upon UCF-Crime, Yuan \etal~\cite{yuan2024towards} introduce new modalities by annotating the event content within 1,854 of its videos. 
Concurrently, Wu~\etal~\cite{wu2024toward} propose a new task termed Video Anomaly Retrieval (VAR), which aims to retrieve relevant segments from long, untrimmed videos through cross-modal queries. 
In~\cite{wu2024toward}, Wu~\etal also construct the first VAR benchmark, UCFCrime-AR, by annotating textual descriptions for UCF-Crime videos.
Our work focuses on the Video Anomaly Retrieval task. To address the data scarcity caused by long-tail distributions, we propose a large-scale Synthetic Video-Text Anomaly benchmark (SVTA) generated by generative models.

\noindent\textbf{Video-Text Retrieval.}
Video-Text Retrieval (VTR) is a multi-modal task that aims to retrieve the most relevant video/text content based on a textual/video query. VTR is a widely studied yet challenging task, whose core lies in achieving cross-modal fusion between different modalities and gaining a deep understanding of the temporal dynamics within videos.
Early VTR works~\cite{torabi2016learning, kiros2014unifying, yu2018joint, kaufman2017temporal} extract video~\cite{li2023fine} and text features offline, aligning the different modalities by designing complex cross-modal fusion strategies. These methods are constrained by the extracted unimodal features. To address this limitation, recent research has focused on end-to-end video-text retrieval~\cite{miech2020end, lei2021less, bain2021frozen, wang2022align}. 
With the advancement of pre-trained models~\cite{radford2021learning, li2022blip, ALBEF, xvlm}, particularly the advent of CLIP~\cite{radford2021learning}, video-text retrieval has become predominantly driven by pre-trained models, focusing on both zero-shot retrieval and fine-tuning tasks~\cite{gan2023cnvid, wu2023empirical, wang2023rtq}.
In this paper, we select two CLIP-based VTR methods, CLIP4Clip~\cite{luo2022clip4clip} and X-CLIP~\cite{ma2022x}, along with a SOTA multimodal large model, GRAM~\cite{cicchetti2025gramian}, as our baselines. Extensive experiments validate the challenging nature of our proposed SVTA dataset and the effectiveness of SVTA in training robust cross-modal retrieval models.

\begin{figure*}[t]
\centering
\vspace{-.15in}
\subfloat[Overall distribution of normal and anomaly videos.
\label{fig:ratio}
]
{\begin{minipage}[b]{0.33\linewidth}
\includegraphics[width=1\linewidth]{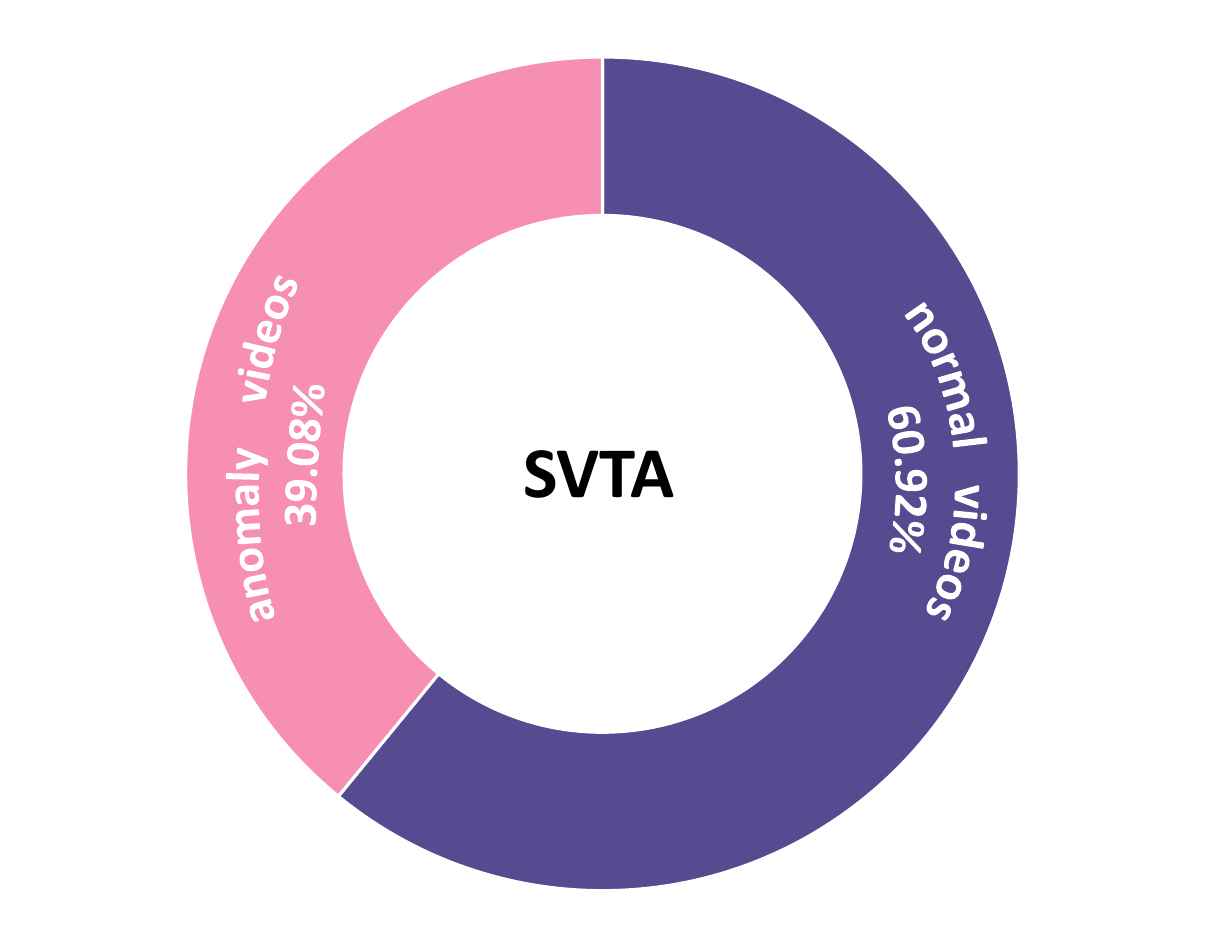}
\end{minipage}
}
\hspace{-1em}
\subfloat[Distribution of normal types.
\label{fig:normal}
]
{\begin{minipage}[b]{0.33\linewidth}
\includegraphics[width=1\linewidth]{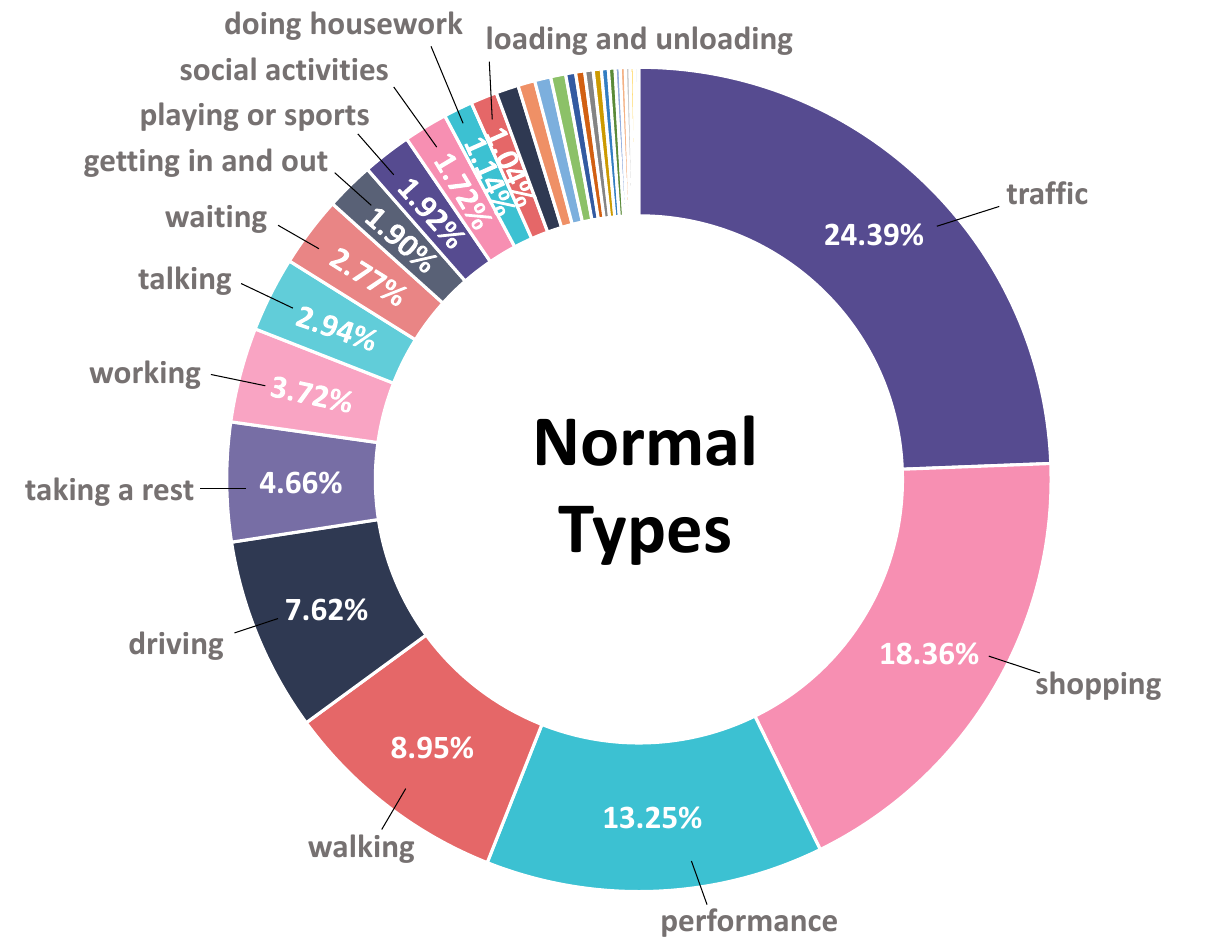}
\end{minipage}
}
\hspace{-1em}
\subfloat[Distribution of anomaly types.
\label{fig:anomaly}]
{
\begin{minipage}[b]{0.33\linewidth}
\includegraphics[width=1\linewidth]{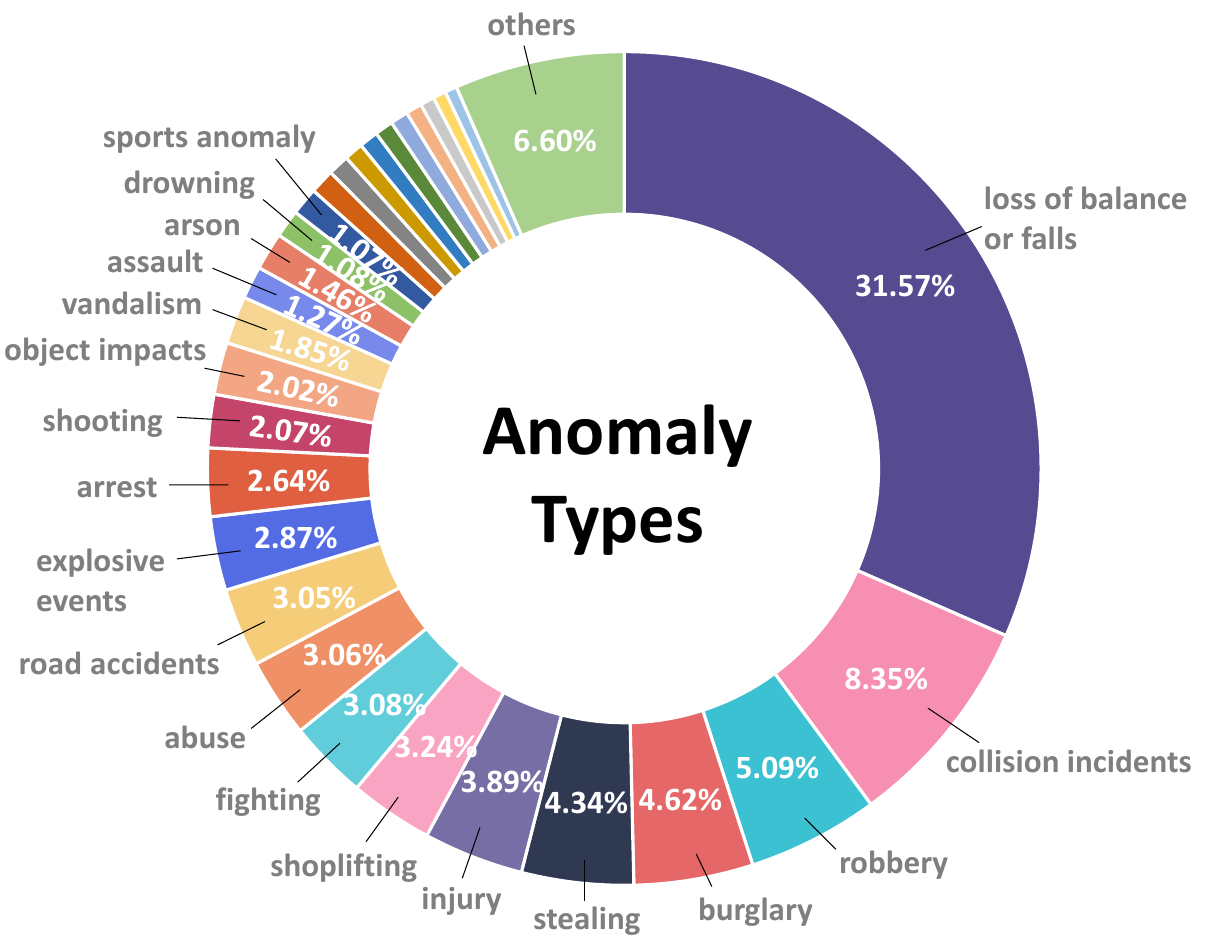}
\end{minipage}}
\vspace{-.1in}
\caption{{\bf Dataset Statistics.} An overview of the SVTA attribute annotations, including the distribution of (a) normal and anomaly videos, (b) normal categories, and (c) anomaly categories. 
}
\label{fig:pie}
\vspace{-.15in}
\end{figure*}

\begin{figure}[t]
\centering
\subfloat[Distributions of word frequency.
\label{fig:frequency}
]
{\begin{minipage}[b]{0.5\linewidth}
\includegraphics[width=1\linewidth]{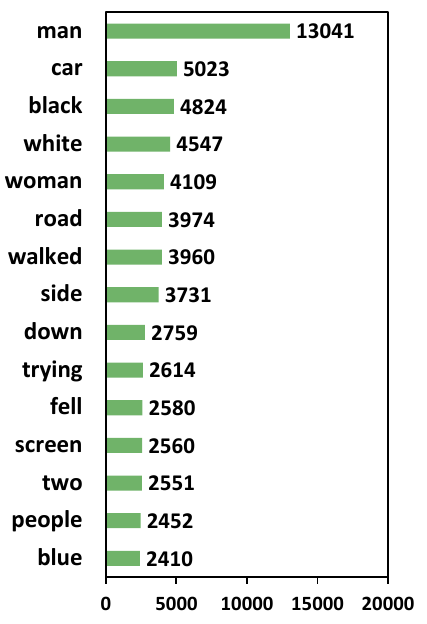}
\end{minipage}
}
\hspace{-1em}
\subfloat[Sentence Length.
\label{fig:number}]
{
\begin{minipage}[b]{0.5\linewidth}
\includegraphics[width=1\linewidth]{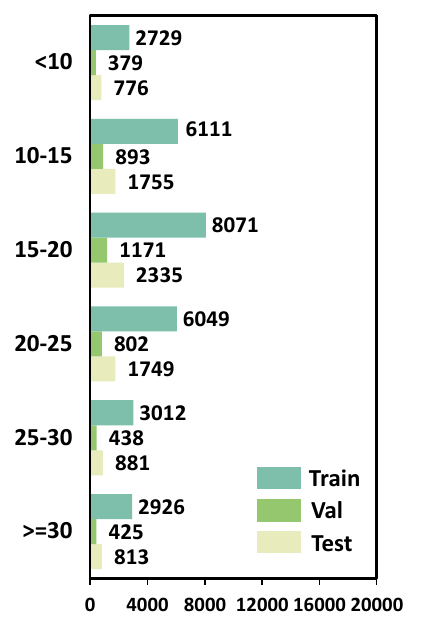}
\end{minipage}}
\vspace{-.1in}
\caption{{\bf Text statistics:} (a) statistical histogram distributions of word frequency on SVTA, and (b) the sentence length in the video caption of the SVTA dataset. 
}
\label{fig:text}
\vspace{-.15in}
\end{figure}

%% file: sec/3_bench.tex
\section{Benchmark}
Data scarcity caused by the long-tail nature of real-world anomalies and privacy constraints hinder large-scale anomaly data collection from real-world scenarios. Fortunately, recent advances in Large Language Models (LLMs) and video foundation models enable high-quality text and video generation. Inspired by this, we propose to leverage LLMs and video generative models to create a large-scale cross-modal video anomaly retrieval benchmark. In Figure \ref{fig:pipe}, we provide an overview of our Synthetic Video-Text Anomaly (SVTA) benchmark construction.

\subsection{Diverse Video Descriptions Acquisition}
We adopt UCA~\cite{yuan2024towards}, UCFCrime-AR~\cite{wu2024toward}, and OOPS!~\cite{epstein2020oops} as text sources due to their rich diversity of normal and anomalous video descriptions. We retain all video descriptions from UCA and UCFCrime-AR, while filtering ambiguous texts from OOPS! by removing descriptions containing ``don’t know". 
To further enhance anomaly diversity, we collect 23 unique anomaly categories from UBnormal~\cite{acsintoae2022ubnormal} and NTU RGB+D 120~\cite{liu2019ntu}: throw, running, falling, fighting, sleeping, crawling, having a seizure, laying down, dancing, stealing, rotating 360 degrees, shuffling, walking injured, walking drunk, stumbling walk, people and car accident, car crash, running injured, fire, smoke, jaywalking, driving outside lane, and jumping.
Using LLMs (Qwen3~\cite{qwen3}, Seed-Thinking-v1.5~\cite{seed2025seed}, and DeepSeek R1~\cite{deepseekai2025} here), we expand these 23 anomaly categories into $23 \times 50$ diverse textual descriptions (see Figure \ref{fig:pipe}). We manually filter unreasonable or low-quality samples, retaining 50 descriptions per category with variations in age, gender, cultural context, environment, and camera angles. 
Finally, we acquire 41,315 video descriptions, comprising 16,145 for normal and 25,170 for anomalous ones. The text statistics are shown in Figure \ref{fig:text}.

\begin{table*}
\centering
  \footnotesize
   \vspace{-.15in}
  \begin{tabular}{l|ccccc|ccccc}
    \toprule
    \multirow{2}{*}{Method} & & & T2V & & & & & V2T & & \\
    & R@1$\uparrow$ & R@5$\uparrow$ & R@10$\uparrow$ & MdR$\downarrow$ & MnR$\downarrow$ & R@1$\uparrow$ & R@5$\uparrow$ & R@10$\uparrow$ & MdR$\downarrow$ & MnR$\downarrow$ \\
    \midrule
    CLIP4Clip-MeanP~\cite{luo2022clip4clip} & 54.0 & 81.7 & 88.9 & 1.0 & 8.8 & 55.8 & 82.5 & 89.4 & 1.0 & 7.9 \\
    CLIP4Clip-seqLSTM~\cite{luo2022clip4clip} & 53.9 & 81.7 & 88.7 & 1.0 & 8.7 & 55.7 & 82.4 & 89.4 & 1.0 & 7.8 \\
    CLIP4Clip-seqTransf~\cite{luo2022clip4clip} & 55.4 & 82.6 & 89.4 & 1.0 & 7.9 & 55.7 & 82.9 & 89.7 & 1.0 & 7.6 \\
    CLIP4Clip-tightTransf~\cite{luo2022clip4clip} & 46.3 & 75.6 & 84.7 & 2.0 & 15.3 & 46.9 & 76.2 & 85.2 & 2.0 & 16.3 \\
    X-CLIP (ViT-B/32)~\cite{ma2022x} & 52.9 & 79.9 & 88.1 & 1.0 & 9.0 & 52.9 & 80.2 & 87.9 & 1.0 & 9.4 \\
    X-CLIP (ViT-B/16)~\cite{ma2022x} & 55.8 & 82.2 & 89.6 & 1.0 & 8.0 & 56.2 & 82.1 & 89.4 & 1.0 & 8.1 \\
    GRAM~\cite{cicchetti2025gramian} & 57.3 & 82.0 & 88.7 & 1.0 & 130.5 & 56.5 & 81.6 & 88.3 & 1.0 & 137.9 \\
    \bottomrule
  \end{tabular}
  \vspace{-.1in}
  \caption{Multimodal text-to-video (T2V) and video-to-text (V2T) retrieval results in terms of Recall Rate (R@1, R@5, R@10), Median Rank (MdR), and Mean Rank (MnR) on SVTA.}
  \label{tab:svta}
  \vspace{-.15in}
\end{table*}

\subsection{High-Quality and Efficient Video Generation}
We employ a state-of-the-art open-source text-to-video model, \ie, Wan2.1~\cite{wan2025}, to generate high-quality videos aligned with textual prompts. The synthetic videos and their corresponding texts form matched cross-modal pairs. 
To optimize generation efficiency and quality, we adopt three strategies specifically:
1) Leveraging Wan2.1-T2V-14B~\cite{wan2025} with FP8 quantization via DiffSynth-Studio~\cite{DiffSynth-Studio};
2) Accelerating video generation with TeaCache~\cite{liu2025teacache} (threshold=0.08);
3) Setting video parameters: 33 total frames, 15 fps, $480 \times 832$ resolution ($H \times W$), quality $=6$.
All generations use a fixed random seed (42) for reproducibility. 
Human inspection reveals that the content of generated videos is realistic and reasonable to some extent, though there is minor noise inherited from text-to-video generative models. Our experiments show that such minor noise is tolerable for deep learning methods.

\subsection{Attribute Annotation}
Following the aforementioned steps, we obtained {\bf 41,315 video-text pairs (1.36 million frames)}. A subset of these pairs includes anomaly category labels from their original data sources, while the remaining anomaly and normal samples lack explicit category annotations. To address this, we leverage the text comprehension capability of LLM (Qwen3~\cite{qwen3} here) by designing tailored instructions (see Figure \ref{fig:pipe}). By inputting video descriptions, we guide the LLM to annotate behavioral categories for the text.
We manually evaluate $10\%$ of the annotated results through random sampling, and the accuracy exceeds $97\%$. Since the LLM annotates each text independently, the same kind of anomaly or normal activity might receive distinct category names, \eg, ``drone crash'', ``drone falling'', and ``drone malfunction''. To unify the annotations, we first cluster all category names using K-means and then manually refine the clustering results. This process yields {\bf 30 normal activities} and {\bf 68 anomaly events}, as illustrated in Figure \ref{fig:pie}.
Specifically, 30 normal activities are categorized into:
fundamental motion patterns, \eg, walking and standing, and contextually consistent actions, \eg, playing soccer on a sports field.
68 anomaly types are systematically organized into four taxonomies:
1) Biomechanical anomalies, \eg, falling and involuntary lying down.
2) Contextual inconsistencies, \eg, running in a laboratory setting.
3) Social norm violations, \eg, hitting and pushing.
4) Environmental anomalies, \eg, natural disasters and object collapses.

\subsection{Dataset Analysis}
Through the above steps, we have successfully constructed the Synthetic Video-Text Anomaly Benchmark (SVTA), which is the first large-scale dataset designed for cross-modal anomaly retrieval.
In Table \ref{tab:dataset}, we compare SVTA with existing video anomaly detection and retrieval datasets across dimensions such as data modality, the number of videos and texts, annotation granularity, anomaly diversity, normal-to-anomaly sample ratios, and data sources. 
SVTA exhibits the following distinctive characteristics:
\begin{itemize}[leftmargin=*]
\item {\bf Extensive anomaly coverage:}  Compared to existing anomaly detection datasets (22 anomalies in UBnormal~\cite{acsintoae2022ubnormal}) and anomaly retrieval datasets (13 anomalies in UCFCrime-AR~\cite{wu2024toward} and UCA~\cite{yuan2024towards}), SVTA contains {\bf 68 anomaly categories} that comprehensively cover all anomalies from these three benchmarks. This scale better aligns with real-world anomaly diversity.
\item {\bf High-quality video samples:} Unlike real CCTV footage suffering from poor illumination and blurry artifacts, synthetic videos from SVTA maintain rational content logic, photorealistic quality, and aesthetic appeal.
\item {\bf Diversity:} We collect textual descriptions from three distinct sources and expand anomaly categories via two complementary benchmarks, ensuring diversity in data provenance and anomaly types. Our LLM-powered anomaly-to-description conversion further enriches variations in character portrayals, environmental contexts, and camera perspectives. All these strategies result in video-text sample diversity.
\item {\bf Large-scale video-text pairs:} The dataset comprises 41k aligned video-text pairs, providing sufficient data for deep cross-modal models to learn discriminative unimodal features and enhanced cross-modal interactions.
\item {\bf Rich annotations:} Beyond matched text descriptions, SVTA provides explicit normal or anomaly category labels as well as attribute annotations for each video, supporting further fine-grained exploration of anomaly behavior analysis.
\item {\bf Less privacy concerns:} As all videos are synthetically generated without involving real human subjects, SVTA inherently mitigates ethical concerns and privacy issues. 
\end{itemize}

\noindent{\bf Discussion: Key Advantages of SVTA.}
Existing anomaly retrieval datasets derive annotations from pre-existing anomaly detection datasets, inheriting their limitations. 
For instance, UCFCrime-AR~\cite{wu2024toward} and UCA~\cite{yuan2024towards} rely on UCF-Crime~\cite{sultani2018real}, \ie, a video anomaly detection dataset with only 13 anomaly types and 1,900 video samples, resulting in constrained anomaly diversity and dataset scale. 
In contrast, SVTA expands anomaly types via LLM-driven annotation and synthesizes high-quality videos using generative models, enabling theoretically unbounded generation of diverse anomaly samples. 
In this work, we generate 41,315 video-text pairs (1.36M frames) covering 30 normal activities and 68 anomaly events, significantly surpassing UCFCrime-AR and UCA.
Furthermore, the construction process of SVTA avoids privacy constraints inherent to real-world datasets, mitigating ethical and legal concerns.

\begin{table*}
\centering
  \footnotesize
   \vspace{-.15in}
  \begin{tabular}{l|ccccc|ccccc}
    \toprule
    \multirow{2}{*}{Method} & & & T2V & & & & & V2T & & \\
    & R@1$\uparrow$ & R@5$\uparrow$ & R@10$\uparrow$ & MdR$\downarrow$ & MnR$\downarrow$ & R@1$\uparrow$ & R@5$\uparrow$ & R@10$\uparrow$ & MdR$\downarrow$ & MnR$\downarrow$ \\
    \midrule
    CLIP4Clip-MeanP~\cite{luo2022clip4clip} & 23.6 & 50.0 & 63.0 & 5.5 & 15.7 & 16.7 & 39.5 & 54.1 & 9.0 & 22.6 \\
    CLIP4Clip-seqLSTM~\cite{luo2022clip4clip} & 22.9 & 49.0 & 64.4 & 6.0 & 16.0 & 18.4 & 36.1 & 52.4 & 10.0 & 23.5 \\
    CLIP4Clip-seqTransf~\cite{luo2022clip4clip} & 24.0 & 47.6 & 64.0 & 6.0 & 16.1 & 17.7 & 36.4 & 51.0 & 10.0 & 22.4 \\
    CLIP4Clip-tightTransf~\cite{luo2022clip4clip} & 16.8 & 41.4 & 53.4 & 8.0 & 32.9 & 14.3 & 34.0 & 49.0 & 12.0 & 39.4 \\
    X-CLIP (ViT-B/32)~\cite{ma2022x} & 24.0 & 49.7 & 63.4 & 6.0 & 16.4 & 17.7 & 36.4 & 52.7 & 9.0 & 22.7 \\
    X-CLIP (ViT-B/16)~\cite{ma2022x} & 27.4 & 53.1 & 67.8 & 5.0 & 14.0 & 20.4 & 44.6 & 59.5 & 7.0 & 19.6 \\
    GRAM~\cite{cicchetti2025gramian} & 34.5 & 60.7 & 70.7 & 3.0 & 17.8 & 32.4 & 57.2 & 68.6 & 4.0 & 26.3 \\
    \bottomrule
  \end{tabular}
  \vspace{-.1in}
  \caption{Zero-shot multimodal text-to-video (T2V) and video-to-text (V2T) retrieval results in terms of Recall Rate (R@1, R@5, R@10), Median Rank (MdR), and Mean Rank (MnR) on UCFCrime-AR.}
  \label{tab:ucf}
\end{table*}

\begin{table*}
\centering
  \footnotesize
  \vspace{-.1in}
  \begin{tabular}{l|ccccc|ccccc}
    \toprule
    \multirow{2}{*}{Method} & & & T2V & & & & & V2T & & \\
    & R@1$\uparrow$ & R@5$\uparrow$ & R@10$\uparrow$ & MdR$\downarrow$ & MnR$\downarrow$ & R@1$\uparrow$ & R@5$\uparrow$ & R@10$\uparrow$ & MdR$\downarrow$ & MnR$\downarrow$ \\
    \midrule
    CLIP4Clip-MeanP~\cite{luo2022clip4clip} & 16.1 & 35.9 & 45.6 & 14.0 & 112.4 & 14.0 & 31.2 & 40.6 & 19.0 & 127.6 \\
    CLIP4Clip-seqLSTM~\cite{luo2022clip4clip} & 15.5 & 35.5 & 45.2 & 14.0 & 114.0 & 12.8 & 30.5 & 39.5 & 20.0 & 129.5 \\
    CLIP4Clip-seqTransf~\cite{luo2022clip4clip} & 16.0 & 35.5 & 45.5 & 14.0 & 114.4 & 13.0 & 30.6 & 40.6 & 19.0 & 127.4 \\
    CLIP4Clip-tightTransf~\cite{luo2022clip4clip} & 11.8 & 27.7 & 36.8 & 24.0 & 219.3 & 9.5 & 24.9 & 33.3 & 29.0 & 230.8 \\
    X-CLIP (ViT-B/32)~\cite{ma2022x} & 15.7 & 35.6 & 46.9 & 13.0 & 108.0 & 14.1 & 32.8 & 41.8 & 17.0 & 110.9 \\
    X-CLIP (ViT-B/16)~\cite{ma2022x} & 18.9 & 41.8 & 52.2 & 9.0 & 84.4 & 16.5 & 36.8 & 47.0 & 12.0 & 91.9 \\
    GRAM~\cite{cicchetti2025gramian} & 18.6 & 39.0 & 48.1 & 12.0 & 542.2 & 19.0 & 39.1 & 49.1 & 11.0 & 551.0 \\
    \bottomrule
  \end{tabular}
    \vspace{-.1in}
    \caption{Zero-shot multimodal text-to-video (T2V) and video-to-text (V2T) retrieval results in terms of Recall Rate (R@1, R@5, R@10), Median Rank (MdR), and Mean Rank (MnR) on OOPS!.}
  \label{tab:oops}
  \vspace{-.15in}
\end{table*}

\begin{figure*}[t]
  \centering
  \vspace{-.15in}
  \includegraphics[width=0.98\linewidth]{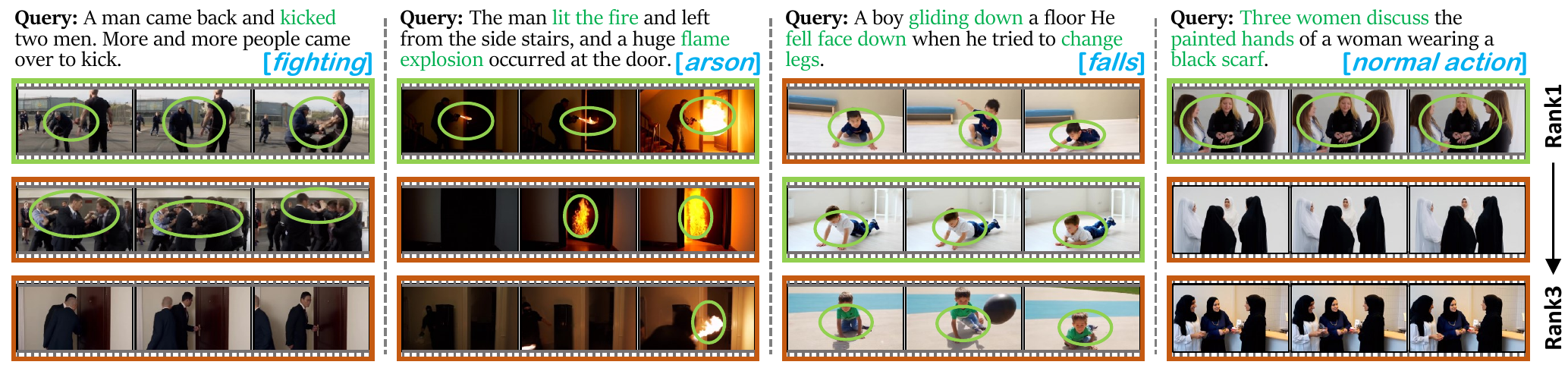}
  \vspace{-.1in}
  \caption{Some retrieved examples of GRAM on SVTA. We visualize top 3 retrieved videos (\textcolor[rgb]{0.57,0.81,0.31}{green}: correct; \textcolor[rgb]{0.77,0.35,0.06}{orange}: incorrect).}
  \label{fig:retrieval}
  \vspace{-.15in}
\end{figure*}

%% file: sec/5_experiment.tex
\section{Experiment}

\subsection{Datasets and Evaluation Metrics}
\noindent {\bf Dataset Splits.}
SVTA is evenly partitioned into training, validation, and test sets with a ratio of $7:1:2$. Specifically, for any action category in the dataset, we randomly allocate $7/10$ video-text pairs to the training set, $1/10$ to the validation set, and the remaining $2/10$ to the test set. The training set contains $28,898$ video-text pairs, while the validation and test sets comprise $4,108$ and $8,309$ pairs, respectively. All experiments strictly adhere to this split.

\noindent {\bf Evaluation Metrics.}
We adopt standard retrieval metrics to assess anomaly retrieval performance: Recall@K (R@K, higher is better), Median Rank (MdR, lower is better), and Mean Rank (MnR, lower is better).
During inference, given a query sample (text/video), all gallery samples (videos/texts) are ranked based on their relevance to the query.
R@K (Recall at rank K) calculates the percentage of test samples where the correct result appears within the top-K retrieved results. We report R@1, R@5, and R@10.
MdR (Median Rank) denotes the median position of the true match in the ranked list.
MnR (Mean Rank) represents the arithmetic average of all correct result rankings.
In this work, we report both text-to-video retrieval results and their complementary video-to-text counterparts for comprehensive reference.

\subsection{Video Anomaly Retrieval Results}

\noindent {\bf Baselines.}
We choose three widely-used video-text retrieval methods, \ie, CLIP4Clip~\cite{luo2022clip4clip}, X-CLIP~\cite{ma2022x}, and GRAM~\cite{cicchetti2025gramian} as our baseline models.
CLIP4Clip~\cite{luo2022clip4clip} transfers knowledge from the image-language pretraining model CLIP~\cite{radford2021learning} to video-text retrieval in an end-to-end manner. CLIP4Clip proposes three types of video-text similarity calculators, \ie, parameter-free, sequential, and tight types.
X-CLIP~\cite{ma2022x}, another CLIP-based framework, is a multi-grained contrastive model for video-text retrieval. Through multi-grained contrastive learning and Attention Over Similarity Matrix, X-CLIP enhances focus on essential frame-word interactions while mitigating the influence of irrelevant frames and words on retrieval outcomes.
GRAM~\cite{cicchetti2025gramian}, a multi-modal model of 1B parameters (text, audio, video), introduces a novel similarity metric for multi-modal representation learning and alignment. It ensures geometric consistency across $n$ modalities while delivering richer and more informative semantic representations.
We comprehensively evaluate SVTA using these three baseline methods.
We train CLIP4Clip~\cite{luo2022clip4clip} and X-CLIP on SVTA for 5 epochs each and fine-tune GRAM for 5 epochs. All experiments are conducted on four NVIDIA GeForce RTX 3090 GPUs, with parameter configurations strictly following their official implementations.

\noindent {\bf Quantitative evaluation.}
As shown in Table \ref{tab:svta}, we show multimodal text-to-video (T2V) and video-to-text (V2T) retrieval results on SVTA.
Methods with comparable parameter scales, namely CLIP4Clip~\cite{luo2022clip4clip} and X-CLIP~\cite{ma2022x}, show competitive performance on SVTA. 
For CLIP4Clip~\cite{luo2022clip4clip}, we report results from four distinct similarity calculator variants. Among them, CLIP4Clip-seqTransf achieves the best text-to-video anomaly retrieval performance, with R@1$=55.4 \%$, R@5$=82.6 \%$, and R@10$=89.4 \%$.
X-CLIP~\cite{luo2022clip4clip} initialized with CLIP (ViT-B/32) slightly underperforms relative to all CLIP4Clip~\cite{luo2022clip4clip} variants except CLIP4Clip-tightTransf. 
Switching to CLIP (ViT-B/16) initialization yields a significant $2.9 \%$ R@1 improvement in text-to-video retrieval, aligning with general video-text retrieval performance trends.
The larger-scale state-of-the-art (SOTA) video-text model GRAM~\cite{cicchetti2025gramian} achieves the best results in both video-to-text and text-to-video retrieval tasks, attaining R@1 of $57.3 \%$ and $56.5 \%$, respectively. 
It should be noted that GRAM employs additional re-ranking techniques, resulting in significantly higher (\ie, worse) Mean Rank (MnR) values compared to CLIP4Clip and X-CLIP.
These results collectively underscore the challenging nature of SVTA as a benchmark dataset.

\noindent {\bf Cross-domain generalization.}
We further investigate the potential of SVTA to train robust cross-modal anomaly retrieval models. Specifically, we conduct domain generalization experiments on two real-world datasets: UCFCrime-AR~\cite{wu2024toward} and OOPS!~\cite{epstein2020oops}. Models trained on SVTA are directly evaluated on these datasets without additional training or fine-tuning.
As shown in Table \ref{tab:ucf}, GRAM~\cite{cicchetti2025gramian} achieves R@1 of $34.5 \%$ (T2V) and $32.4 \%$ (V2T) on UCFCrime-AR~\cite{wu2024toward}, substantially outperforming CLIP4Clip~\cite{luo2022clip4clip} and X-CLIP~\cite{ma2022x} in zero-shot cross-modal retrieval. 
Similarly, Table \ref{tab:oops} presents zero-shot multimodal retrieval results on a filtered version of OOPS!. Notably, our test set differs from the original OOPS! benchmark. We exclude videos lacking complete anomaly descriptions and retain only one matched text per video, resulting in 3,468 curated video-text pairs.
On the T2V task, X-CLIP (ViT-B/16)~\cite{ma2022x} and GRAM~\cite{cicchetti2025gramian} achieve R@1 of $18.9 \%$ and $18.6 \%$, respectively, surpassing other baselines.

\noindent {\bf Qualitative Results.} In Figure~\ref{fig:retrieval}, we show the ranking list of GRAM on our SVTA dataset. We could observe that our dataset is still challenging to prevailing cross-modality models. Some fine-grained actions are not well captured by the existing approaches, which provide a competitive test base for the future exploration from our community. 

%% file: sec/6_conclusion.tex
\section{Conclusion}
We present SVTA in this work, the first large-scale synthetic benchmark for video anomaly retrieval using natural language queries. By leveraging large language models and text to video generative models, SVTA overcomes the data scarcity and privacy limitations of real-world anomaly datasets. SVTA covers 68 anomalous and 30 normal activity categories across 41,315 high-quality video-text pairs, offering a diverse and privacy-compliant evaluation benchmark. Experimental results using the SOTA text-video baselines validate the practical value of SVTA for benchmarking cross-modal anomaly retrieval systems.